\documentclass[a4paper,fleqn]{cas-sc}

\usepackage[authoryear]{natbib}
\usepackage{graphicx}
\usepackage{amsmath}
\usepackage{amsfonts}
\usepackage{float}
\usepackage{placeins}
\usepackage{etoolbox}
\usepackage{booktabs}
\usepackage{colortbl}
\usepackage[table]{xcolor}
\usepackage{tabularx}
\usepackage{ragged2e}
\usepackage{microtype}
\usepackage{bm}
\usepackage{silence}

\definecolor{lafnocdl}{HTML}{DCEBFF}
\definecolor{lafnol}{HTML}{E6F4EA}
\definecolor{lightyellow}{rgb}{1.0, 1.0, 0.8}
\begin{document}
\let\WriteBookmarks\relax
\def\floatpagepagefraction{1}
\def\textpagefraction{.001}

\shorttitle{LAFNO for CT-to-PSMA PET Synthesis}

\shortauthors{Bhaskara et~al.}

\title[mode=title]{Lesion-Aware Adaptive Fourier Neural Operator for CT-to-PSMA PET Synthesis in Prostate Cancer}

\author[1]{Rashmi Bhaskara}

\author[1]{Waleed M. Almutairi}

\author[1]{Matthew Gopaulchan}

\author[1]{Maram Musaad Alqurashi}

\author[2]{Francis Asamoah}

\author[3]{Alex Ocana}

\author[4]{Clinton D. Bahler}

\author[1,2]{Oluwaseyi M. Oderinde\cormark[1]}

\ead{ooderind@purdue.edu}

\affiliation[1]{
    organization={School of Health Sciences, Purdue University},
    city={West Lafayette},
    postcode={47907},
    state={IN},
    country={USA}
}

\affiliation[2]{
    organization={Department of Radiology, Indiana University School of Medicine},
    city={Indianapolis},
    postcode={46202},
    state={IN},
    country={USA}
}

\affiliation[3]{
    organization={Department of Urology, Indiana University School of Medicine},
    city={Indianapolis},
    postcode={46202},
    state={IN},
    country={USA}
}

\affiliation[4]{
    organization={Department of Radiation Oncology, Indiana University School of Medicine},
    city={Indianapolis},
    postcode={46202},
    state={IN},
    country={USA}
}

\cortext[1]{Corresponding author. Oluwaseyi M. Oderinde, PhD, Assistant Professor in Medical Physics. Email: ooderind@purdue.edu}

\begin{abstract}
Deep learning models that synthesize PET from CT or MRI can reduce patient dose and scanner demand, but are typically optimized with global losses such as L1 or mean squared error (MSE) that treat all voxels similarly. In whole-body PSMA-PET, tumor voxels occupy only a small fraction of the volume, yet carry the clinically relevant activity signal; as a result, models can achieve high structural similarity index measure (SSIM) and peak signal-to-noise ratio (PSNR) while still underestimating lesion activity or failing to preserve tumor-specific structure. Radiomics provides biologically meaningful descriptors of tumor intensity and texture, but direct radiomics conditioning is time-consuming because it requires feature extraction from delineated lesion regions. We propose LAFNO, a Lesion-Aware Adaptive Fourier Neural Operator for CT-to-PSMA-PET synthesis that replaces high-dimensional radiomics conditioning with two efficient CT-derived proxy channels. Motivated by radiomics analysis of PSMA-avid tumor-core and peritumoral regions, LAFNO uses a contrast proxy for local density variation and a disorder proxy for local texture heterogeneity, both injected into the model bottleneck. LAFNO combines whole-volume reconstruction with lesion-level total lesion activity (TLA), tumor-core contrast, and peritumoral supervision. We evaluated LAFNO against four baseline architectures on the TCIA PSMA-PET-CT-Lesions dataset. LAFNO remained competitive on whole-volume image quality, achieving SSIM of 0.960 and 0.938 for $^{18}$F- and $^{68}$Ga-PSMA, respectively, while reducing per-patient TLA error to 48.3\% and 64.0\% for $^{18}$F- and $^{68}$Ga-PSMA, respectively, and achieving the highest tumor-core radiomics reproducibility across all feature classes for both tracers. Peritumoral reproducibility remained tracer-dependent, indicating that biological fidelity in synthetic PSMA-PET remains challenging.
\end{abstract}

\begin{highlights}
\item We propose LAFNO for lesion-aware CT-to-PSMA PET synthesis in prostate cancer.
\item CT-derived contrast and disorder proxy channels are used to condition the AFNO bottleneck.
\item Lesion-aware losses improve tumor-core and peritumoral fidelity beyond whole-volume metrics.
\end{highlights}


\maketitle

\section{Introduction}
\label{sec:introduction}

Prostate cancer (PCa) is the most diagnosed male malignancy in Western countries \cite{maurer2016current}. In the United States, it accounts for 31\% of all male cancer cases, with more than 35,000 deaths projected in 2026 \cite{jemal2026cancer}. Despite major therapeutic advances, many patients, particularly those with high-risk disease, still experience recurrence or progression \cite{bouchelouche2016psma, tonry2020clinical}. Accurate imaging therefore plays a critical role in diagnosis, staging, restaging, biochemical recurrence detection, and treatment planning \cite{bouchelouche2016psma}.

Prostate-specific membrane antigen (PSMA) is a transmembrane glycoprotein expressed by prostatic epithelial cells, with higher expression in malignant prostate tissue than in benign or normal tissue. PSMA expression increases with Gleason grade and tumor stage, making it an important biomarker in both hormone-sensitive and castration-resistant prostate cancer \cite{kim2023evaluation}. Because biochemical changes often precede anatomical changes \cite{reddy2010immuno}, conventional imaging modalities such as computed tomography (CT) and magnetic resonance imaging (MRI) can be limited for early detection and disease characterization. PSMA PET/CT has therefore become increasingly important because it provides improved sensitivity for staging, restaging, and detection of metastatic disease compared with conventional imaging \cite{siva2020expanding, tsechelidis2022psma, caldarella2024role, dondi2024prognostic, ishibashi2023prognostic}.

Despite its diagnostic value, PSMA PET/CT remains expensive and unevenly accessible \cite{subramanian2023complex, smith2025prostate}. Limited tracer availability, scanner demand, insurance coverage, and cost can restrict access, particularly in underserved populations. These barriers are clinically important because delayed or incomplete imaging may affect staging accuracy, recurrence detection, and treatment planning \cite{subramanian2023complex}. Therefore, alternative imaging strategies that reduce dependence on radiotracer administration while preserving clinically relevant tumor information could improve accessibility and broaden the clinical impact of PSMA-targeted imaging.

Recent advances in machine learning and deep learning have enabled cross-modality medical image synthesis, including PET synthesis from anatomical imaging \cite{dayarathna2024deep}. Convolutional neural networks, especially U-Net-based encoder--decoder architectures, have been widely used for medical image-to-image translation because they preserve spatial structure through skip connections \cite{ronneberger2015u, emami2020frea}. Generative models, including GANs, Pix2Pix, CycleGAN, and diffusion models, have also been explored for multimodal medical image synthesis \cite{ghafari2022generation, chai2025synthesizing, jung2024generation}. 

However, CT-to-PET synthesis in prostate cancer remains challenging because CT has limited soft-tissue contrast, particularly for small structures and subtle lesions \cite{abdollahi2026microenvironment}. These lesions occupy only a tiny fraction of the voxels in a whole-body scan \cite{abtahi2026fine}. As a result, models trained with global reconstruction objectives and evaluated mainly using whole-volume metrics such as structural similarity index measure (SSIM), peak signal-to-noise ratio (PSNR), and mean absolute error (MAE) can become biased toward healthy background tissue and may omit or underestimate small but clinically important diagnostic hotspots \cite{abtahi2026fine}. In addition, the mapping from CT to PET is fundamentally one-to-many: multiple plausible uptake patterns may correspond to the same CT appearance, especially for subtle lesions and physiologic tracer activity \cite{mahdi2026ct}. These limitations highlight the need for synthetic PET models that move beyond global image similarity and incorporate tumor-relevant biological and anatomical signatures.


Radiomics provides a useful bridge between anatomical imaging and tumor biology because it converts standard medical images into high-dimensional, mineable quantitative features that can describe tumor intensity, shape, and texture \cite{lambin2012radiomics, gillies2016radiomics}. These features have been explored as non-invasive imaging biomarkers across multiple cancers, including lung cancer, glioblastoma, and prostate cancer \cite{lambin2012radiomics, gillies2016radiomics, almutairi2026metabolic}. Recent work has also shown that radiomics can be used as a conditioning signal for tumor synthesis. Kim et al. proposed a radiomics-conditioned tumor-generation framework that uses a GAN-based model to generate tumor masks and a diffusion-based model to generate tumor texture conditioned on user-specified radiomics features such as size, shape, and texture \cite{kim2025tumor}. This supports the idea that radiomics features can provide biologically meaningful conditioning for generative medical imaging models. However, conventional radiomics extraction can be time-consuming because it requires computing numerous handcrafted features from delineated lesion regions. The extraction also depends on accurate lesion delineation, and manual or semi-automated segmentation can be time-consuming, operator-dependent, and sensitive to inter-observer variability \cite{traverso2018repeatability, gillies2016radiomics}.

Building on this idea, we present a \textbf{Lesion-aware Adaptive Fourier Neural Operator (LAFNO)} for CT-to-PSMA-PET synthesis. Rather than using radiomics features directly as model inputs, we use radiomics analysis as a discovery step to identify repeatable CT patterns across all annotated PSMA-avid lesions in The Cancer Imaging Archive (TCIA) PSMA-PET-CT-Lesions dataset~\cite{jeblick2026whole}. We then replace high-dimensional radiomics conditioning with two computationally simple CT-derived proxy channels that approximate these patterns: a contrast proxy for local density variation and a disorder proxy for local texture heterogeneity. These proxies are used to condition our model. In this way, LAFNO preserves the biological motivation of radiomics conditioning while avoiding radiomics extraction and lesion segmentation during inference.

Importantly, these proxy patterns are not defined only from the tumor core. Our analysis also includes the peritumoral region because tumor progression and therapeutic resistance are increasingly understood to depend not only on tumor cell biology, but also on continuous signaling with the surrounding tumor microenvironment (TME), including immune cells, cancer-associated fibroblasts, stromal cells, vasculature, and secreted signaling molecules \cite{wang2021tumor, liu2025value}. The peritumoral region, the interface between tumor and adjacent normal tissue, remains less studied than the tumor core \cite{koca2024peritumoral, zhang2023peritumor}, despite evidence that it can be transcriptionally and phenotypically distinct from both tumor and healthy tissue \cite{koca2024peritumoral}. In prostate cancer, pronounced intratumoral heterogeneity limits the reliability of single-region assessment \cite{yadav2018intratumor}. Incorporating peritumoral radiomics features has been shown to complement intratumoral features and improve diagnostic and classification performance beyond intratumoral analysis alone \cite{algohary2020combination, zhou2025integrating}. Notably, the diagnostic contribution of peritumoral features may depend on their spatial distance from the tumor boundary \cite{liu2025value}. Therefore, our radiomics analysis explicitly includes the peritumoral region, and motivates the disorder proxy, which approximates local texture heterogeneity around tumor-adjacent tissue. Together with the contrast proxy, this allows LAFNO to condition PET synthesis on both tumor-core density variation and peritumoral texture patterns.

In summary, to address the limitations of global CT-to-PET synthesis models, we use radiomics analysis to identify repeatable tumor-core and peritumoral CT patterns and translate these patterns into efficient CT-derived proxy channels. Based on this design, we propose \textbf{LAFNO}, a lesion-aware CT-to-PSMA-PET synthesis framework that combines proxy-conditioned image synthesis with lesion-level and peritumoral supervision. The main contributions are summarized as follows:

\begin{itemize}
    \item We introduce CT-derived proxy conditioning to guide PET synthesis using radiomics-motivated tumor-core and peritumoral patterns.

    \item We propose a lesion-aware objective that supervises total lesion activity and penalizes errors in the tumor-adjacent peritumoral region.

    \item We evaluate synthetic PET beyond whole volume similarity using lesion-level uptake metrics and tumor/peritumoral radiomics reproducibility.
\end{itemize}

\section{Materials and Methods}
\label{sec:methods}

\subsection{CT Radiomics Analysis of Tumor-Core and Peritumoral Shells}
\label{sec:radiomics}

\begin{figure}
	\centering
	\includegraphics[width=.75\textwidth]{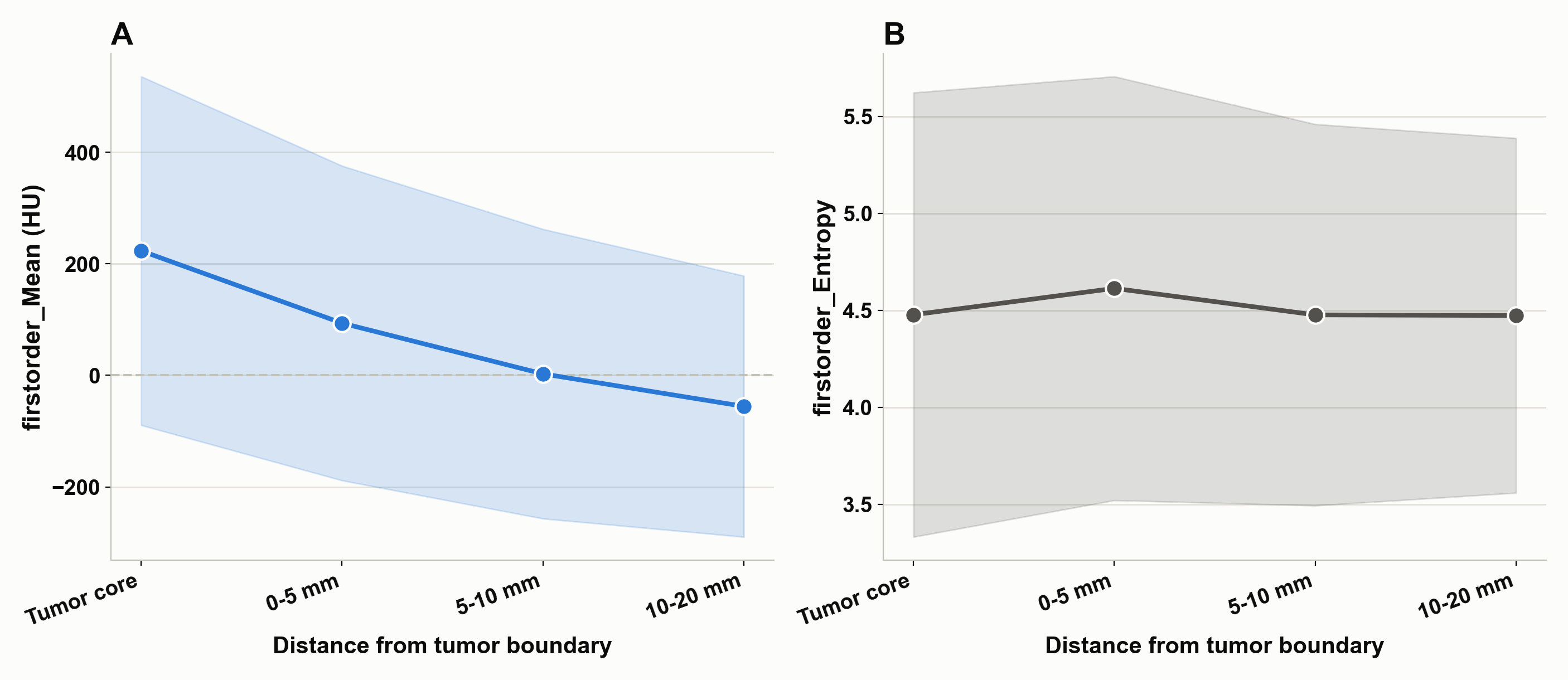}
	\caption{CT radiomics profiles from the tumor core to peritumoral shells. \textbf{(A)} \texttt{firstorder\_Mean} decreases outward. \textbf{(B)} \texttt{firstorder\_Entropy} increases from tumor core to peritumoral tissue and remains relatively stable.}
	\label{fig:combined}
\end{figure}

We analyzed whether CT radiomics features vary systematically from the tumor core outward. We studied 10,209 connected-component PSMA-avid tumor regions across 480 patients. For each tumor region derived from the ground-truth mask, radiomics features were extracted independently from the tumor core and the peritumoral shells at increasing distances from the tumor boundary: 0--5\,mm, 5--10\,mm, and 10--20\,mm. Shells were constructed using a Euclidean distance transform from the tumor boundary, with tumor voxels excluded from every shell. Adjacent regions were compared using paired Wilcoxon signed-rank tests with false discovery rate correction, and the analysis was repeated after patient-level aggregation to reduce the influence of patients with many lesions.

Two spatial trends were observed (Figure~\ref{fig:combined}). First, CT attenuation decreased monotonically from the tumor core to the outer peritumoral shells. This was reflected by \texttt{firstorder\_Mean} and \texttt{firstorder\_Median}, which quantify the average and central CT intensity within each region, respectively, and by \texttt{ngtdm\_Contrast}, which measures local gray-level contrast relative to neighboring voxels \cite{zwanenburg2016image}. The \texttt{firstorder\_Mean} trend was significant across all adjacent shell comparisons at the lesion level and remained consistent after patient-level aggregation, indicating a robust core-to-periphery attenuation gradient.

Second, CT texture heterogeneity increased when moving from the tumor core into the peritumoral region. This was captured by \texttt{firstorder\_Entropy}, which measures randomness or heterogeneity in the intensity distribution \cite{zwanenburg2016image}. Entropy increased significantly on leaving the tumor core at both lesion and patient levels. At the lesion level, entropy peaked in the 0--5\,mm shell, suggesting increased heterogeneity near the tumor boundary. At the patient level, entropy increased from the tumor core to the peritumoral tissue and then remained relatively stable across the outer shell, indicating a tumor-to-peritumoral increase in CT texture heterogeneity.

Together, these findings show a smooth core-to-periphery attenuation gradient and a tumor-to-peritumoral increase in texture. These two CT signatures directly motivate the contrast and disorder proxy channels described in Section~\ref{sec:proxies}.

\subsection{Differentiable CT Proxy Channels}
\label{sec:proxies}

\begin{figure}
	\centering
	\includegraphics[width=.65\textwidth]{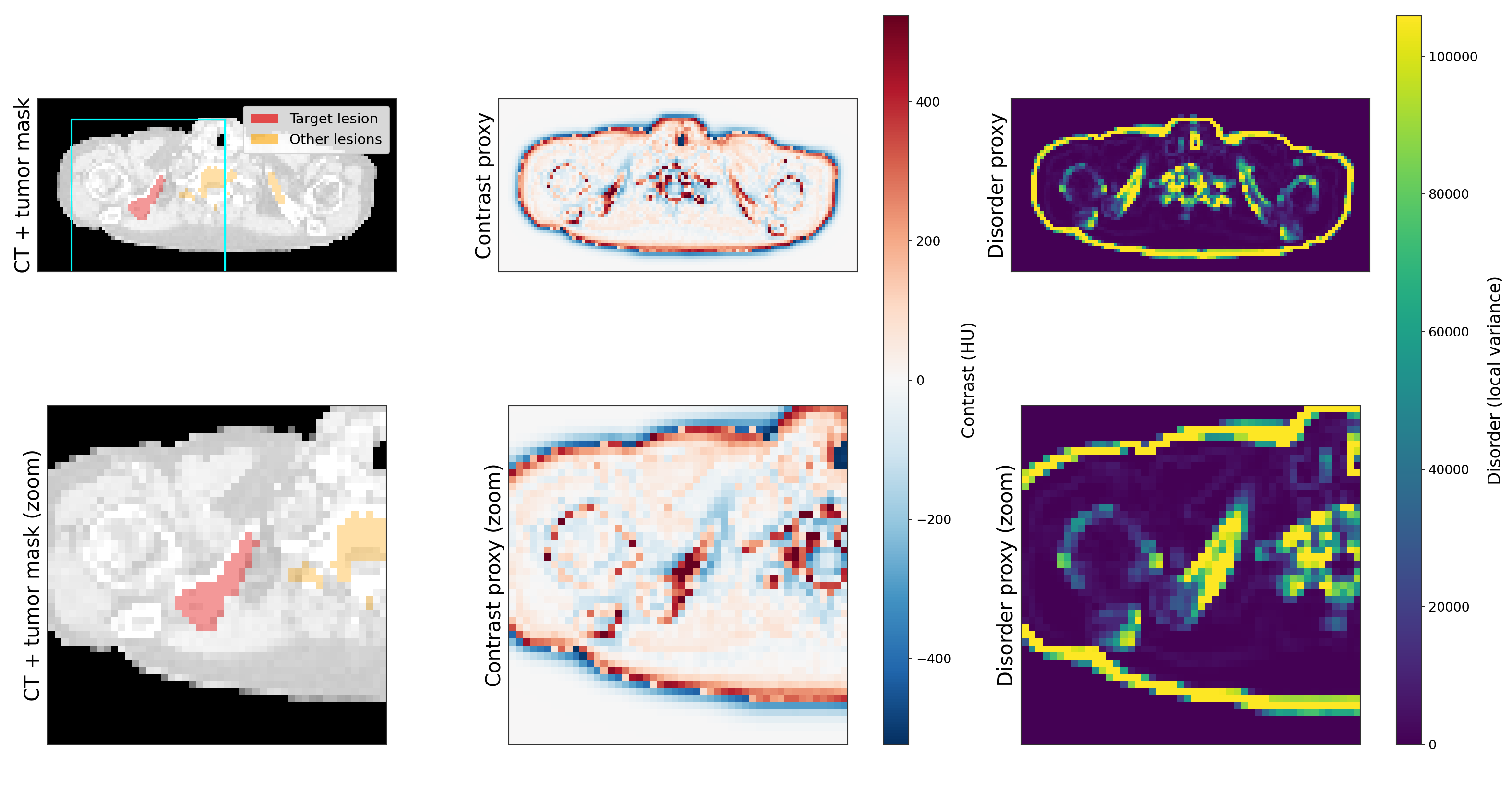}
	\caption{Representative CT-derived proxy maps showing the CT image, contrast proxy, and disorder proxy. Zoomed panels highlight the lesion neighborhood.}
	\label{fig:proxies}
\end{figure}

To encode the two CT signatures, we introduce two differentiable proxy filters, contrast and disorder, that can be computed directly from CT without requiring a ground-truth tumor mask. Representative proxy maps are shown in Figure~\ref{fig:proxies}. The contrast proxy approximates the core-to-periphery attenuation gradient using a Gaussian residual filter:

\begin{equation}
C(\mathbf{x}) = \text{CT}(\mathbf{x}) - G_{\sigma} * \text{CT}(\mathbf{x}), \qquad \sigma = 5\,\text{mm}
\label{eq:contrast}
\end{equation}

where $\mathbf{x}$ denotes the voxel location, $\text{CT}(\mathbf{x})$ is the CT intensity at that voxel, $C(\mathbf{x})$ is the resulting contrast proxy value, $G_{\sigma}$ is a 3D Gaussian smoothing kernel with physical scale $\sigma$, and $*$ denotes convolution. The Gaussian smoothing is implemented using separable 1D convolutions. The scale $\sigma = 5\,\text{mm}$ was chosen to match the spatial scale of the observed \texttt{firstorder\_Mean} decay.

The disorder proxy captures tumor-adjacent tissue heterogeneity via local variance over a sliding window:

\begin{equation}
D(\mathbf{x}) = \overline{\text{CT}^2}_{w}(\mathbf{x}) - \left(\overline{\text{CT}}_{w}(\mathbf{x})\right)^2, \qquad w = 7.5\,\text{mm}
\label{eq:disorder}
\end{equation}

where $D(\mathbf{x})$ is the disorder proxy value at voxel $\mathbf{x}$, $w$ is the physical width of the local cubic window, $\overline{\text{CT}}_{w}(\mathbf{x})$ is the local mean CT intensity within that window, and $\overline{\text{CT}^2}_{w}(\mathbf{x})$ is the local mean of squared CT intensities. The local means are implemented using uniform-filter convolution. The window size $w = 7.5\,\text{mm}$ was chosen empirically to capture local tumor--peritumoral texture transitions, while avoiding excessive sensitivity to voxel-level noise.

 Comparison with the radial radiomics analysis showed that the contrast and disorder proxies capture the observed density and texture trends. Both proxies are computed on the full native-resolution CT volume and used together to highlight tumor-like CT patterns and their interaction with adjacent tissue. In this way, the proxy channels provide the model with an approximate spatial map of regions that may contain tumor-relevant behavior. Their integration into the model is described in Section~\ref{sec:arch}.

\subsection{LAFNO Architecture}
\label{sec:arch}

\begin{figure}
	\centering
	\includegraphics[width=.75\textwidth]{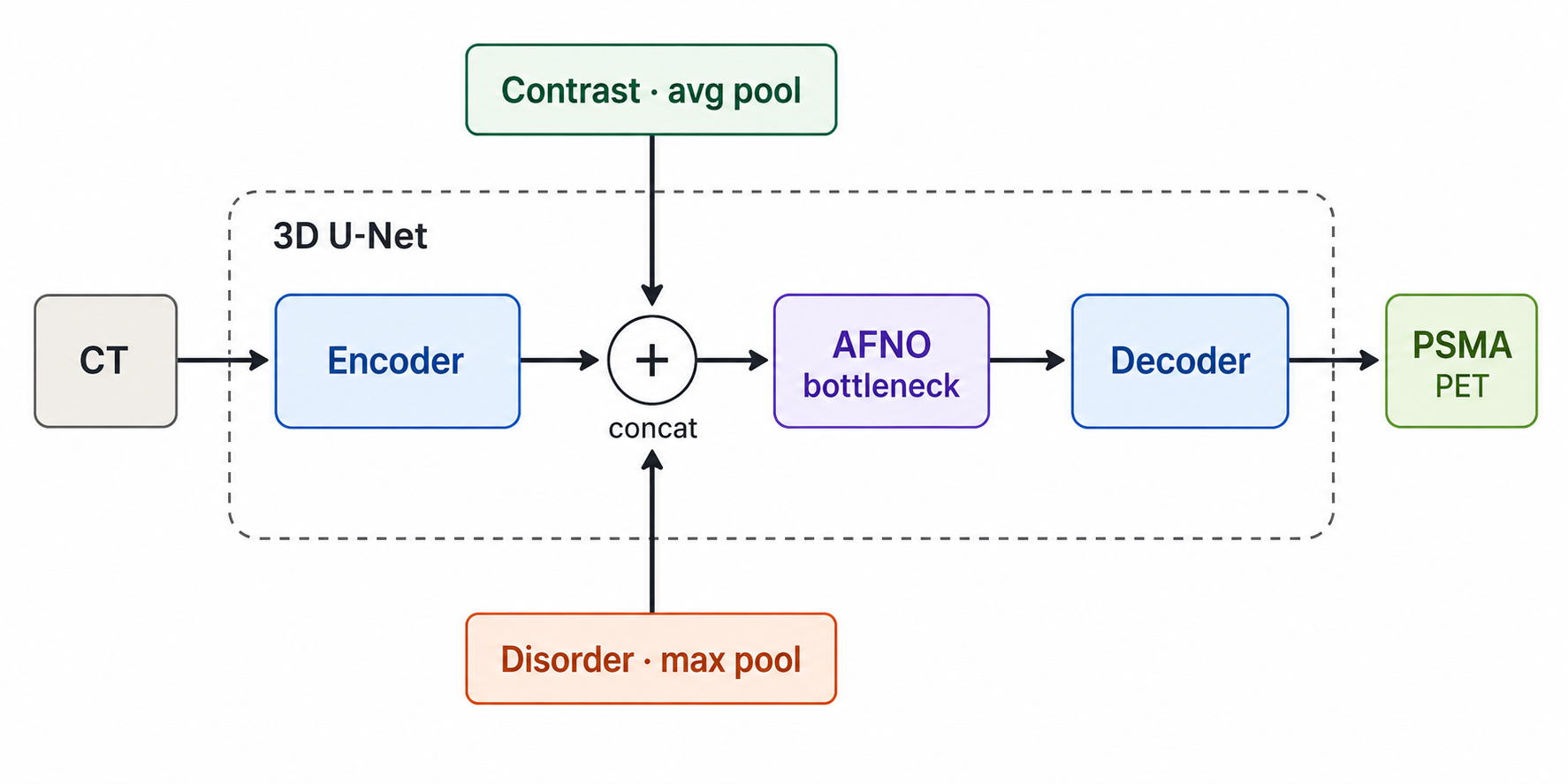}
	\caption{Overview of the proposed LAFNO architecture with CT proxy conditioning and lesion-aware training supervision.}
	\label{fig:architecture}
\end{figure}

LAFNO is a 3D U-Net with an Adaptive Fourier Neural Operator (AFNO) bottleneck~\cite{guibas2021adaptive}. The encoder applies three stride-2 convolution blocks (Conv3D + BatchNorm + LeakyReLU), mapping a $64^3$ CT patch to an $8^3$ representation with 256 channels. The decoder mirrors the encoder with transposed convolution upsampling and skip connections, ending with a Sigmoid output.

At the $8^3$ bottleneck, four AFNO blocks perform spectral channel mixing followed by a residual MLP. Before the first AFNO block, the two CT proxy channels are injected. The contrast proxy is average-pooled from $64^3$ to $8^3$ to preserve its smooth monotonic gradient, while the disorder proxy is max-pooled to preserve its localized peak character. The pooled proxies are concatenated with the 256-channel bottleneck features and projected back to 256 channels:

\begin{equation}
\mathbf{b}' = W_{258\to256}\!\left[\,
\mathbf{b} \;\|\;
\text{AvgPool}_8\bigl(C\bigr) \;\|\;
\text{MaxPool}_8\bigl(D\bigr)
\,\right]
\label{eq:proj}
\end{equation}

where $\mathbf{b} \in \mathbb{R}^{256 \times 8^3}$ are the bottleneck features, $W_{258\to256}$ is a learned $1{\times}1{\times}1$ convolution, and $\|$ denotes channel concatenation. The projected features $\mathbf{b}'$ are then passed through the four AFNO blocks. Figure~\ref{fig:architecture} shows the overall architecture.

\subsection{Lesion-Aware Loss}
\label{sec:loss}

The training objective combines whole-volume PET reconstruction with lesion-level activity preservation and peritumoral supervision:


\begin{equation}
\mathcal{L} = \mathcal{L}_{\text{L1}} +
\lambda_{\text{TLA}}\,\mathcal{L}_{\text{TLA}} +
\lambda_{c}\,\mathcal{L}_{c} +
\lambda_{\text{peri}}\mathcal{L}_{\text{peri}}
\label{eq:loss}
\end{equation}

Here, $\mathcal{L}_{\text{L1}}$ is the standard voxel-wise L1 loss. Because background voxels vastly outnumber lesion voxels, a voxel-wise loss alone tends to under-constrain small, high-uptake regions. The lesion-level term $\mathcal{L}_{\text{TLA}}$ addresses this by supervising the summed activity of each connected component independently, so that small lesions are not dominated by larger lesions or by surrounding background tissue. For a set of lesions $\mathcal{K}$ in a patient volume, the lesion-wise activity loss is defined as

\begin{equation}
\mathcal{L}_{\text{TLA}} =
\frac{1}{|\mathcal{K}|}
\sum_{k \in \mathcal{K}}
\frac{\left|\log\left(1+\widehat{A}_k\right) -
\log\left(1+A_k\right)\right|}
{\log\left(1 + N_k \cdot SUV_{\max}\right)} ,
\label{eq:tla}
\end{equation}

where $N_k$ is the number of voxels in lesion $k$, $\mathrm{SUV}_{\max}$ is the global standardized uptake value (SUV) normalization ceiling, and

\begin{equation}
\widehat{A}_k = \sum_{i \in k}\hat{s}_i ,
\qquad
A_k = \sum_{i \in k}s_i .
\label{eq:summed_activity}
\end{equation}

Here, $\hat{s}_i$ and $s_i$ are the predicted and ground-truth SUV values within the same lesion mask. The logarithmic transform compresses the wide dynamic range of lesion uptake, and the denominator normalizes each lesion term by its maximum possible summed activity, $N_k \cdot SUV_{\max}$. This keeps lesion contributions on a comparable scale and weights lesions by count rather than by absolute activity magnitude. The quantities $\widehat{A}_k$ and $A_k$ are summed SUV values. Physical total lesion activity can be obtained by multiplying by the voxel volume, $\text{TLA}_k = v_{\text{vox}} A_k$. In the loss, we use the summed-SUV form because the predicted and ground-truth activities are computed over the same lesion mask and therefore remain proportional to physical TLA. The objective is minimized when $\widehat{A}_k = A_k$, which corresponds to matching physical TLA for that lesion.

The remaining two lesion-aware terms supervise tumor-core contrast and peritumoral fidelity. The tumor-contrast loss, $\mathcal{L}_{c}$, applies the contrast operation from Equation~\ref{eq:contrast} to the predicted and ground-truth PET volumes in SUV space and averages the absolute difference over tumor-mask voxels. This penalizes mismatch in local tumor-core uptake structure.

The peritumoral loss, $\mathcal{L}_{\text{peri}}$, is computed in the 0--10\,mm ring outside the tumor mask using an exponential distance-weighted L1 penalty. Let $R(\mathbf{x})$ denote the binary peritumoral ring mask and let $d(\mathbf{x})$ be the distance in millimeters from voxel $\mathbf{x}$ to the tumor boundary. The peritumoral weight field is defined as

\begin{equation}
w(\mathbf{x}) =
R(\mathbf{x}) \exp\left(-\frac{d(\mathbf{x})}{\tau}\right),
\qquad \tau = 5\,\text{mm}.
\label{eq:peri_weight}
\end{equation}

The decay parameter was set to $\tau=5\,\text{mm}$, with this value, the weight decreases to half of its boundary value from the tumor boundary. Finally, the peritumoral loss is then computed as

\begin{equation}
\mathcal{L}_{\text{peri}} =
\frac{
\sum_{\mathbf{x}} w(\mathbf{x})
\left|
\hat{y}(\mathbf{x}) - y(\mathbf{x})
\right|
}{
\sum_{\mathbf{x}} w(\mathbf{x}) + \epsilon
},
\qquad \epsilon = 10^{-6},
\label{eq:peri_loss}
\end{equation}

where $\hat{y}(\mathbf{x})$ and $y(\mathbf{x})$ are the predicted and ground-truth PET values. This loss gives higher importance to voxels immediately surrounding the tumor and gradually reduces the penalty for voxels farther from the tumor boundary. As a result, the model is encouraged to preserve PET structure in the tumor-adjacent region without overemphasizing distant background tissue.

Together, these terms encourage the model to preserve lesion activity, tumor-core uptake structure, and peritumoral behavior rather than optimizing whole-volume similarity alone. Loss weights were set to $\lambda_{\text{L1}}=1.0$, $\lambda_{\text{TLA}}=0.05$, $\lambda_{c}=0.02$, and $\lambda_{\text{peri}}=0.05$.

\subsection{Training}
\label{sec:training}

All models were trained patch-wise on $64^3$ crops for 200 epochs using a single NVIDIA A100-40\,GB GPU. We used the Adam optimizer with learning rate $2{\times}10^{-4}$, $\beta_1=0.5$, and $\beta_2=0.999$, followed by linear learning-rate decay from epoch 50. To address tumor sparsity, 70\% of training patches were lesion-centered with random positional jitter, while 30\% were sampled uniformly to match the sliding-window inference distribution. Patients were sampled uniformly before patch extraction.

\section{Experimental Setup and Evaluation}

\begin{figure}
	\centering
	\includegraphics[width=0.9\textwidth]{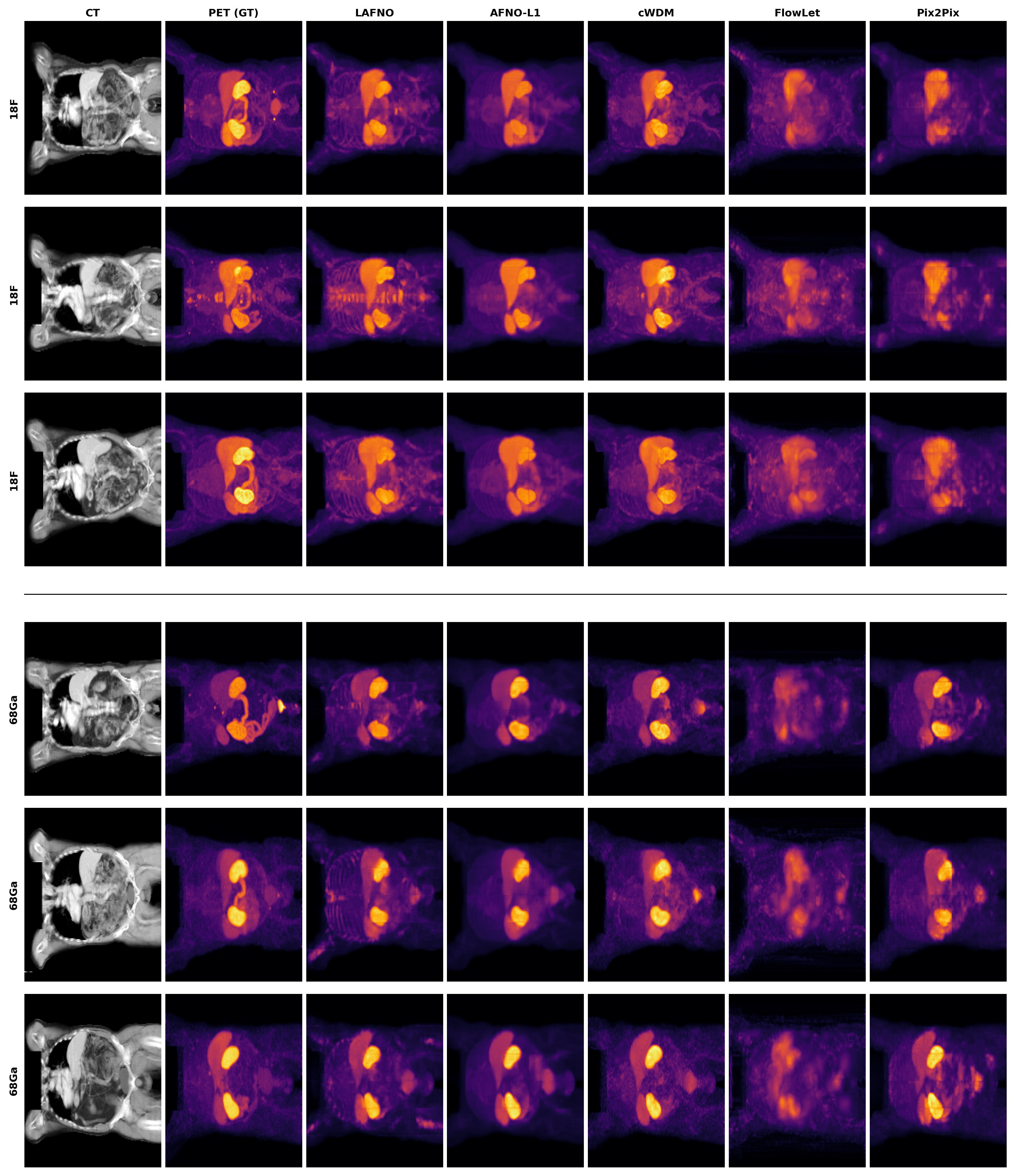}
	\caption{Representative reconstructions from different CT-to-PET synthesis models.}
	\label{fig:mip_examples}
\end{figure}

\subsection{Baselines}
\label{sec:baselines}

We selected one representative baseline from each major CT-to-PET synthesis training strategy: U-Net-style reconstruction, adversarial training, flow matching, and diffusion-based generation.

\textbf{AFNO-L1.}
AFNO-L1 was used as the direct architectural baseline for LAFNO. It follows the same 3D U-Net-style encoder--decoder design with AFNO blocks inserted at the bottleneck, but receives only the CT patch as input and is trained with whole-volume L1 loss. Unlike LAFNO, it does not use CT-derived proxy conditioning or lesion-aware supervision. This baseline isolates the effect of the proposed proxy channels and lesion-aware loss while keeping the core architecture fixed.

\textbf{Pix2Pix.}
Pix2Pix~\cite{isola2017image} was used as a 3D conditional GAN baseline. The generator is a 3D U-Net that maps a $64^3$ CT patch to a synthetic PET patch using encoder--decoder layers with skip connections, and the discriminator follows a PatchGAN design.

\textbf{FlowLet.}
FlowLet~\cite{Danese_2026} was used as a flow-matching baseline. The model operates in 3D Haar wavelet space, where CT and PET volumes are decomposed into low- and high-frequency subbands. A U-Net backbone is trained to predict the flow from CT wavelet coefficients to PET wavelet coefficients, conditioned on the CT wavelet representation and a time embedding. The predicted PET wavelet coefficients are converted back to image space using the inverse wavelet transform.

\textbf{cWDM.}
cWDM~\cite{friedrich2024cwdmconditionalwaveletdiffusion} was used as a conditional diffusion baseline. The model also operates in 3D Haar wavelet space and uses a U-Net with attention blocks. CT wavelet coefficients are used as the conditioning input, and the final PET volume is reconstructed from generated PET wavelet coefficients using the inverse wavelet transform.

\subsection{Dataset and Preprocessing}
\label{sec:data}

We used TCIA PSMA-PET-CT-Lesions dataset~\cite{jeblick2026whole}, including $^{18}$F-PSMA ($n=335$) and $^{68}$Ga-PSMA ($n=204$) patients with annotated tumor masks. CT volumes were clipped to $[-1000,1000]$\,HU and linearly scaled to $[0,1]$. PET volumes were converted to body-weight standardized uptake value (SUV) and log-normalized as

\begin{equation}
\text{PET}(\mathbf{x}) =
\frac{\log(1 + \text{SUV}(\mathbf{x}))}{\log(1 + SUV_{\max})},
\label{eq:petnorm}
\end{equation}

where $\mathbf{x}$ denotes the voxel location and $SUV_{\max}$ is the global cohort SUV ceiling. During evaluation, predictions were converted back to SUV using the inverse transform. The CT-derived proxy channels $C(\mathbf{x})$ and $D(\mathbf{x})$ were computed before resampling and normalized by their respective 99th percentiles.

\subsection{Evaluation Metrics}
\label{sec:evaluation}

Whole-volume image quality was assessed using the Structural Similarity Index Measure (SSIM), Peak Signal-to-Noise Ratio (PSNR), and Mean Absolute Error (MAE). Whole-volume SUVmax and SUVmean percentage errors were reported after converting predictions back from log-normalized to the standardized uptake value (SUV) space.

Lesion-level activity preservation was evaluated using Total Lesion Activity (TLA). Lesions were obtained from the ground-truth tumor mask using 26-connectivity connected-component labeling. For each lesion, physical TLA was computed as the sum of SUV within the lesion mask, multiplied by voxel volume, which is equivalent to SUVmean multiplied by lesion volume. This quantity is proportional to SUV$\cdot$mL TLA because the predicted and ground-truth values are evaluated on the same lesion mask; therefore, the voxel-volume factor cancels in the percentage-error calculation. The percentage error was computed as

\begin{equation}
e_{p,k} =
\frac{\widehat{\text{TLA}}_{p,k} - \text{TLA}_{p,k}}
{\text{TLA}_{p,k}}
\times 100,
\label{eq:tla_error_eval}
\end{equation}

where $\text{TLA}_{p,k}$ and $\widehat{\text{TLA}}_{p,k}$ are the real and predicted TLA values for lesion $k$ in patient $p$. We reported TLA error in two ways. First, lesion-level error was computed by pooling all lesions across all patients and averaging the absolute percentage error. This measures how well the model preserves lesion activity across the full lesion cohort. Second, patient-level error was computed by averaging lesion errors within each patient first and then averaging across patients. We also reported the fraction of lesions within 25\%, 50\%, and 75\% absolute TLA error.

Radiomics reproducibility was evaluated using the Intraclass Correlation Coefficient (ICC) between radiomics features extracted from real and synthesized PET volumes. ICC was used because it measures agreement between real and predicted feature values across the test cohort, rather than only their linear association. Features were extracted from the tumor core and from the 0--10\, mm peritumoral ring outside the tumor mask. For simplicity, radiomics reproducibility was summarized by feature class: first-order features (FO, $n=11$), gray-level co-occurrence matrix features (GLCM, $n=6$), gray-level run-length matrix features (GLRLM, $n=4$), and gray-level size-zone matrix features (GLSZM, $n=4$).

\section{Results}
\label{sec:results}

Figure~\ref{fig:mip_examples} shows representative PET reconstructions from different models. The following sections report the quantitative results.

\subsection{Whole-volume Metrics}

\begin{table}[width=1\linewidth,cols=7,pos=h]
\centering
\caption{Whole-volume image quality across all models. The shaded row is the proposed model; bold marks the best value per column.}
\label{tab:wholevolume}
\scriptsize
\setlength{\tabcolsep}{3pt}
\begin{tabular*}{\tblwidth}{@{\extracolsep{\fill}}lcccccc@{}}
\toprule
& \multicolumn{3}{c}{$^{18}$F-PSMA ($N=47$)} &
  \multicolumn{3}{c}{$^{68}$Ga-PSMA ($N=30$)} \\
\cmidrule(lr){2-4}\cmidrule(lr){5-7}
Model & SSIM $\uparrow$ & PSNR $\uparrow$ & MAE $\downarrow$
      & SSIM $\uparrow$ & PSNR $\uparrow$ & MAE $\downarrow$ \\
\midrule
\rowcolor{lafnocdl}
\textbf{LAFNO (ours)}
& 0.960$\pm$0.014 & 35.98$\pm$1.60 & 0.0038$\pm$0.0012
& 0.938$\pm$0.016 & 35.10$\pm$1.28 & 0.0048$\pm$0.0011 \\
AFNO-L1
& \textbf{0.963$\pm$0.014} & \textbf{36.42$\pm$1.75} & \textbf{0.0036$\pm$0.0012}
& \textbf{0.945$\pm$0.013} & \textbf{35.80$\pm$1.36} & \textbf{0.0044$\pm$0.0009} \\
cWDM
& 0.961$\pm$0.016 & 36.22$\pm$1.80 & 0.0037$\pm$0.0013
& 0.935$\pm$0.015 & 34.96$\pm$1.26 & 0.0049$\pm$0.0010 \\
FlowLet
& 0.934$\pm$0.023 & 33.98$\pm$1.49 & 0.0054$\pm$0.0016
& 0.910$\pm$0.016 & 31.92$\pm$1.02 & 0.0071$\pm$0.0012 \\
Pix2Pix
& 0.944$\pm$0.019 & 33.74$\pm$1.38 & 0.0050$\pm$0.0016
& 0.932$\pm$0.016 & 33.77$\pm$1.30 & 0.0055$\pm$0.0012 \\
\bottomrule
\end{tabular*}
\end{table}

Table~\ref{tab:wholevolume} reports whole-volume SSIM, PSNR, and MAE for both tracer cohorts. AFNO-L1 achieved the best global image-quality metrics on both $^{18}$F-PSMA and $^{68}$Ga-PSMA, while LAFNO showed slightly lower whole-volume fidelity. This trade-off is expected because LAFNO is optimized to preserve tumor-core and peritumoral information rather than maximizing global image similarity alone.

\begin{table}[width=1\linewidth,cols=5,pos=h]
\centering
\caption{Whole-volume SUV error (\% error, mean $\pm$ SD). The shaded row is the proposed model; bold marks the lowest error per column.}
\label{tab:suverror}
\scriptsize
\setlength{\tabcolsep}{4pt}
\begin{tabular*}{\tblwidth}{@{\extracolsep{\fill}}lcccc@{}}
\toprule
& \multicolumn{2}{c}{$^{18}$F-PSMA ($N=47$)} &
  \multicolumn{2}{c}{$^{68}$Ga-PSMA ($N=30$)} \\
\cmidrule(lr){2-3}\cmidrule(lr){4-5}
Model & SUVmax \% err $\downarrow$ & SUVmean \% err $\downarrow$
      & SUVmax \% err $\downarrow$ & SUVmean \% err $\downarrow$ \\
\midrule
\rowcolor{lafnocdl}
\textbf{LAFNO (ours)}
& 55.05$\pm$20.19 & 8.61$\pm$6.29
& 23.60$\pm$17.65 & 13.41$\pm$7.20 \\
AFNO-L1
& 62.98$\pm$18.01 & 10.12$\pm$6.39
& \textbf{18.49$\pm$18.23} & 12.18$\pm$7.01 \\
cWDM
& \textbf{46.14$\pm$24.65} & \textbf{7.24$\pm$7.48}
& 39.84$\pm$35.84 & \textbf{7.52$\pm$5.93} \\
FlowLet
& 67.08$\pm$35.68 & 9.95$\pm$7.05
& 59.60$\pm$13.61 & 13.26$\pm$7.10 \\
Pix2Pix
& 61.71$\pm$18.72 & 19.73$\pm$8.19
& 24.19$\pm$18.31 & 10.84$\pm$6.85 \\
\bottomrule
\end{tabular*}
\end{table}

Table~\ref{tab:suverror} reports whole-volume SUVmax and SUVmean percentage errors. cWDM achieved the lowest SUVmax and SUVmean errors for $^{18}$F-PSMA and the lowest SUVmean error for $^{68}$Ga-PSMA, while AFNO-L1 achieved the lowest SUVmax error for $^{68}$Ga-PSMA. LAFNO remained competitive across global SUV metrics, but was not consistently the best model.

\subsection{Total Lesion Activity (TLA)}

\begin{figure*}
	\centering
	\includegraphics[width=\textwidth]{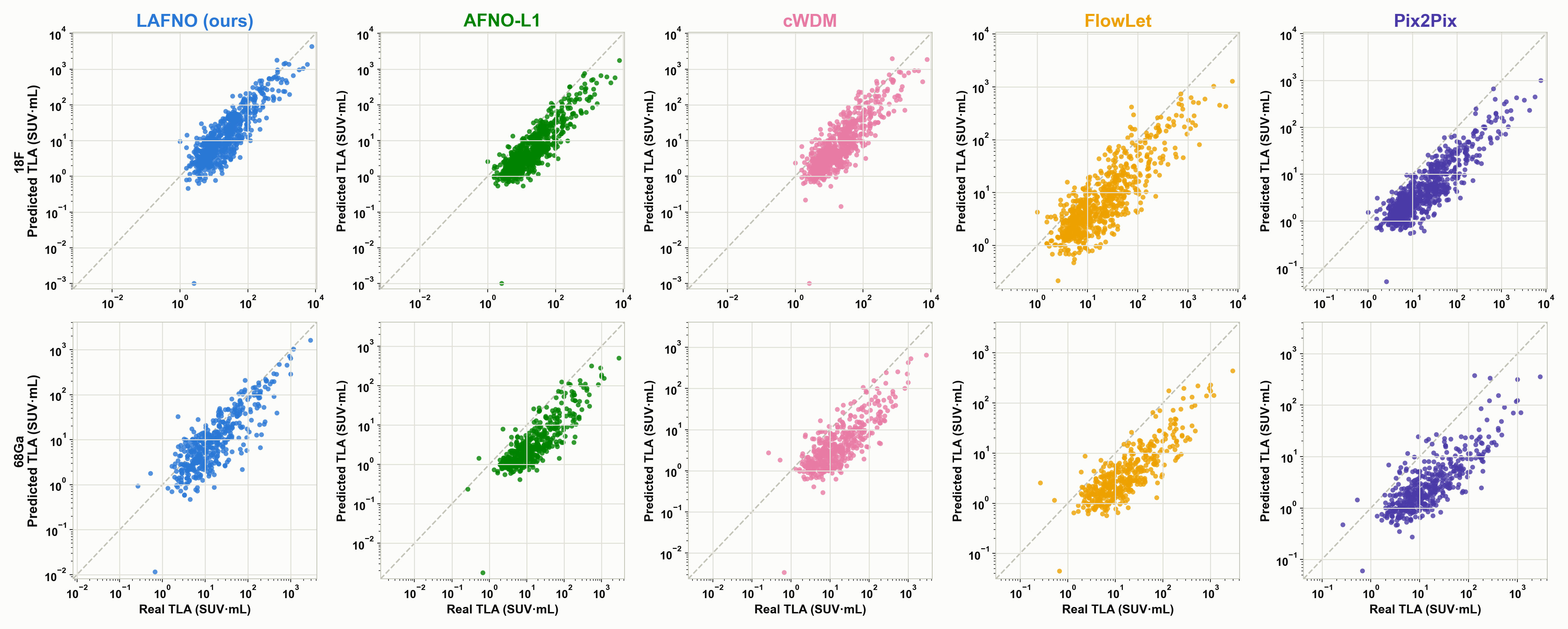}
	\caption{Predicted versus real per-lesion TLA on log--log axes for $^{18}$F-PSMA (top) and $^{68}$Ga-PSMA (bottom). The dashed line indicates perfect agreement; the annotation gives the mean absolute TLA error across all lesions. All models underestimate high-activity lesions, while LAFNO tracks the identity line most closely across the full range of lesion burden.}
	\label{fig:tla_scatter}
\end{figure*}

\begin{table*}[width=1\linewidth,cols=11,pos=h]
\centering
\caption{Lesion-level TLA error. Lesion-level $|$err$|$ pools all lesions; per-patient $|$err$|$ averages within each patient first, then across patients. Within-$X\%$ is the per-patient fraction of lesions with $|$TLA error$| \leq X\%$. The shaded row is the proposed model; bold marks the best value per column.}
\label{tab:tla}
\scriptsize
\setlength{\tabcolsep}{3pt}
\begin{tabular*}{\tblwidth}{@{\extracolsep{\fill}}lccccccccccc@{}}
\toprule
& \multicolumn{5}{c}{$^{18}$F-PSMA} &
  \multicolumn{5}{c}{$^{68}$Ga-PSMA} \\
\cmidrule(lr){2-6}\cmidrule(lr){7-11}
Model & Lesion $|$err$|$ $\downarrow$ & Patient $|$err$|$ $\downarrow$ & $\leq$25\% $\uparrow$ & $\leq$50\% $\uparrow$ & $\leq$75\% $\uparrow$
      & Lesion $|$err$|$ $\downarrow$ & Patient $|$err$|$ $\downarrow$ & $\leq$25\% $\uparrow$ & $\leq$50\% $\uparrow$ & $\leq$75\% $\uparrow$ \\
\midrule
\rowcolor{lafnocdl}
\textbf{LAFNO (ours)}
& \textbf{52.7} & \textbf{48.3} & \textbf{24.0} & \textbf{50.9} & \textbf{81.0}
& \textbf{54.4} & 64.0 & 13.2 & \textbf{31.3} & \textbf{61.4} \\
AFNO-L1
& 66.2 & 62.2 & 9.8 & 25.9 & 65.3
& 73.9 & 71.7 & 4.1 & 17.3 & 43.4 \\
cWDM
& 62.8 & 58.7 & 10.6 & 29.5 & 72.3
& 70.5 & \textbf{63.9} & \textbf{13.6} & 24.5 & 55.6 \\
FlowLet
& 62.3 & 59.2 & 8.3 & 34.3 & 70.9
& 72.6 & 67.6 & 10.8 & 22.8 & 47.6 \\
Pix2Pix
& 70.4 & 69.0 & 4.1 & 15.8 & 52.9
& 78.5 & 71.5 & 10.4 & 22.1 & 39.1 \\
\bottomrule
\end{tabular*}
\end{table*}

Table~\ref{tab:tla} reports lesion-level and patient-level TLA error. For $^{18}$F-PSMA, LAFNO achieved the lowest absolute TLA error and the largest fraction of lesions with TLA error below 25\%, 50\%, and 75\%. The mean absolute patient-level TLA error was reduced to 48.3\%, compared with 58.7--69.0\% for the baseline models. For $^{68}$Ga-PSMA, LAFNO, and cWDM, performance was similar in absolute TLA error, while LAFNO had more lesions per patient with TLA errors below 50\% and 75\%. Figure~\ref{fig:tla_scatter} shows predicted versus real per-lesion TLA on log--log axes. When errors were pooled across all lesions, LAFNO achieved the lowest overall TLA error for both tracers, with 52.7\% error for $^{18}$F-PSMA and 54.4\% error for $^{68}$Ga-PSMA (Table~\ref{tab:tla}). This indicates improved preservation of activity at the lesion level throughout the whole set of lesions. The scatter plots also show that LAFNO more closely follows the identity line, particularly for moderate and high-activity lesions.

\subsection{Radiomics Reproducibility}

\begin{table*}[width=1\linewidth,cols=9,pos=h]
\centering
\caption{Radiomics reproducibility (ICC) between real and synthesized PET, by feature class. Higher is better. The shaded rows are the proposed model; bold marks the best value per column within each region.}
\label{tab:icc}
\scriptsize
\setlength{\tabcolsep}{4pt}
\begin{tabular*}{\tblwidth}{@{\extracolsep{\fill}}lcccccccc@{}}
\toprule
& \multicolumn{4}{c}{$^{18}$F-PSMA} & \multicolumn{4}{c}{$^{68}$Ga-PSMA} \\
\cmidrule(lr){2-5}\cmidrule(lr){6-9}
Model & FO & GLCM & GLRLM & GLSZM & FO & GLCM & GLRLM & GLSZM \\
\midrule
\multicolumn{9}{l}{\textit{Tumor core}} \\
\rowcolor{lafnocdl}
\textbf{LAFNO (ours)}
& \textbf{0.38} & \textbf{0.62} & \textbf{0.78} & \textbf{0.71}
& \textbf{0.27} & \textbf{0.36} & \textbf{0.54} & \textbf{0.47} \\
AFNO-L1 & 0.24 & 0.39 & 0.59 & 0.51 & 0.13 & 0.14 & 0.44 & 0.44 \\
cWDM    & 0.20 & 0.44 & 0.58 & 0.43 & 0.12 & 0.24 & 0.51 & 0.45 \\
FlowLet & 0.17 & 0.29 & 0.40 & 0.35 & 0.17 & 0.30 & 0.48 & 0.44 \\
Pix2Pix & 0.21 & 0.36 & 0.57 & 0.49 & $-$0.03 & 0.08 & 0.34 & 0.37 \\
\midrule
\multicolumn{9}{l}{\textit{Peritumoral ring (0--10\,mm)}} \\
\rowcolor{lafnocdl}
\textbf{LAFNO (ours)}
& \textbf{0.59} & \textbf{0.58} & \textbf{0.85} & \textbf{0.70}
& 0.48 & 0.33 & 0.58 & 0.56 \\
AFNO-L1 & 0.56 & 0.45 & 0.66 & 0.64 & 0.43 & 0.33 & 0.60 & 0.54 \\
cWDM    & 0.55 & 0.49 & 0.73 & 0.61 & \textbf{0.50} & \textbf{0.51} & \textbf{0.72} & 0.49 \\
FlowLet & 0.46 & 0.37 & 0.55 & 0.57 & 0.42 & 0.27 & 0.58 & \textbf{0.58} \\
Pix2Pix & 0.53 & 0.30 & 0.51 & 0.64 & 0.41 & 0.40 & 0.68 & 0.50 \\
\bottomrule
\end{tabular*}

\vspace{0.4em}
\begin{flushleft}
\footnotesize FO = first-order ($n{=}11$); GLCM = gray-level co-occurrence matrix ($n{=}6$); GLRLM = gray-level run-length matrix ($n{=}4$); GLSZM = gray-level size-zone matrix ($n{=}4$).
\end{flushleft}
\end{table*}

\begin{figure*}
	\centering
	\includegraphics[width=0.85\textwidth]{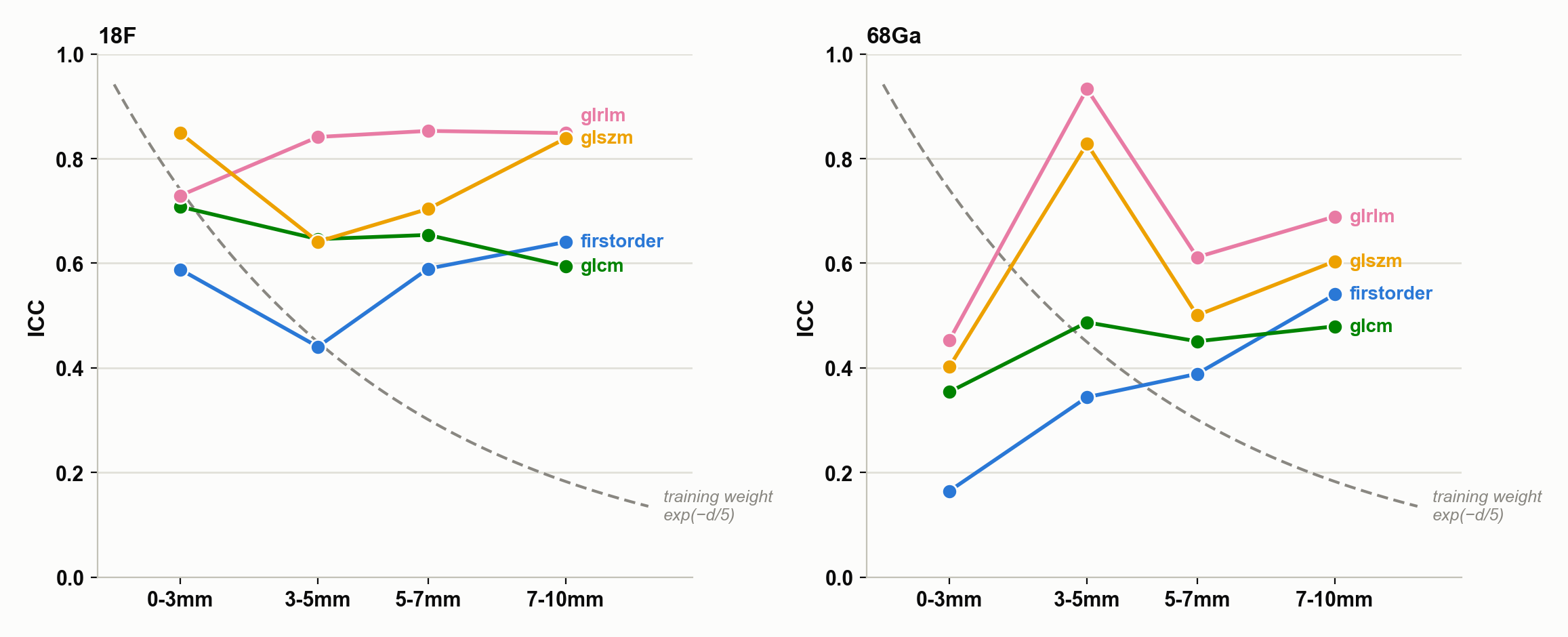}
	\caption{Peritumoral radiomics reproducibility (ICC) by distance band from the tumor boundary, for $^{18}$F-PSMA (left) and $^{68}$Ga-PSMA (right). The dashed curve shows the training loss weight $\exp(-d/5)$.}
	\label{fig:peri_bands}
\end{figure*}

Table~\ref{tab:icc} reports radiomics reproducibility between real and synthesized PET using ICC by feature class. In the tumor core, LAFNO achieved the highest ICC across all feature classes for both $^{18}$F-PSMA and $^{68}$Ga-PSMA, indicating improved preservation of tumor radiomic structure compared with the baseline models. The strongest agreement was observed for texture feature classes such as GLRLM and GLSZM, while first-order features showed lower reproducibility, particularly for $^{68}$Ga-PSMA. This suggests that although LAFNO better preserves tumor texture patterns, recovery of absolute intensity-distribution features remains limited and requires further improvement.

In the 0--10\,mm peritumoral ring, LAFNO achieved the highest ICC across all feature classes for $^{18}$F-PSMA, suggesting improved preservation of tumor-adjacent texture patterns for this tracer. For $^{68}$Ga-PSMA, the peritumoral results were more mixed, with cWDM achieving the highest ICC for FO, GLCM, and GLRLM, and FlowLet achieving the highest GLSZM ICC. Figure~\ref{fig:peri_bands} resolves peritumoral reproducibility by distance
band from the tumor boundary. The two tracers behave differently. For
$^{18}$F-PSMA, the texture classes GLRLM and GLSZM remain high across the full
range, while first-order and GLCM are lowest at the immediate rim and recover
with distance; no single band is best for every class. For $^{68}$Ga-PSMA, all
four classes rise from the 0--3\,mm band outward, and the texture classes peak
sharply at 3--5\,mm before declining. In both tracers, reproducibility at the
immediate rim is not the highest despite the training weight $\exp(-d/5)$ being
largest there, indicating that the region closest to the high-contrast tumor
boundary is the hardest to reproduce.

\section{Ablation Study}

All ablation experiments were conducted on the $^{18}$F-PSMA cohort. We first
compared two formulations of the per-lesion loss term across a range of weights
to establish which better supervises lesion activity
(Section~\ref{sec:abl:tlavstl1}). We then evaluated LAFNO using a cumulative ablation design. Table~\ref{tab:ablation} shows the cumulative ablation study. Starting from AFNO-L1, we added CT-derived proxy conditioning, $\mathcal{L}_{\text{TLA}}$, $\mathcal{L}_{c}$, and $\mathcal{L}_{\text{peri}}$ one step at a time. Each row includes the components from the row above it, so changes between rows show the added effect of each component. The results are discussed in Sections~\ref{sec:abl:proxy}--\ref{sec:abl:peri}.
Throughout, the backbone, training schedule, and test split are held fixed, so
that any difference is attributable to the component under test.

\subsection{Choice of Per-Lesion Loss Formulation}
\label{sec:abl:tlavstl1}

\begin{figure}
	\centering
	\includegraphics[width=.85\textwidth]{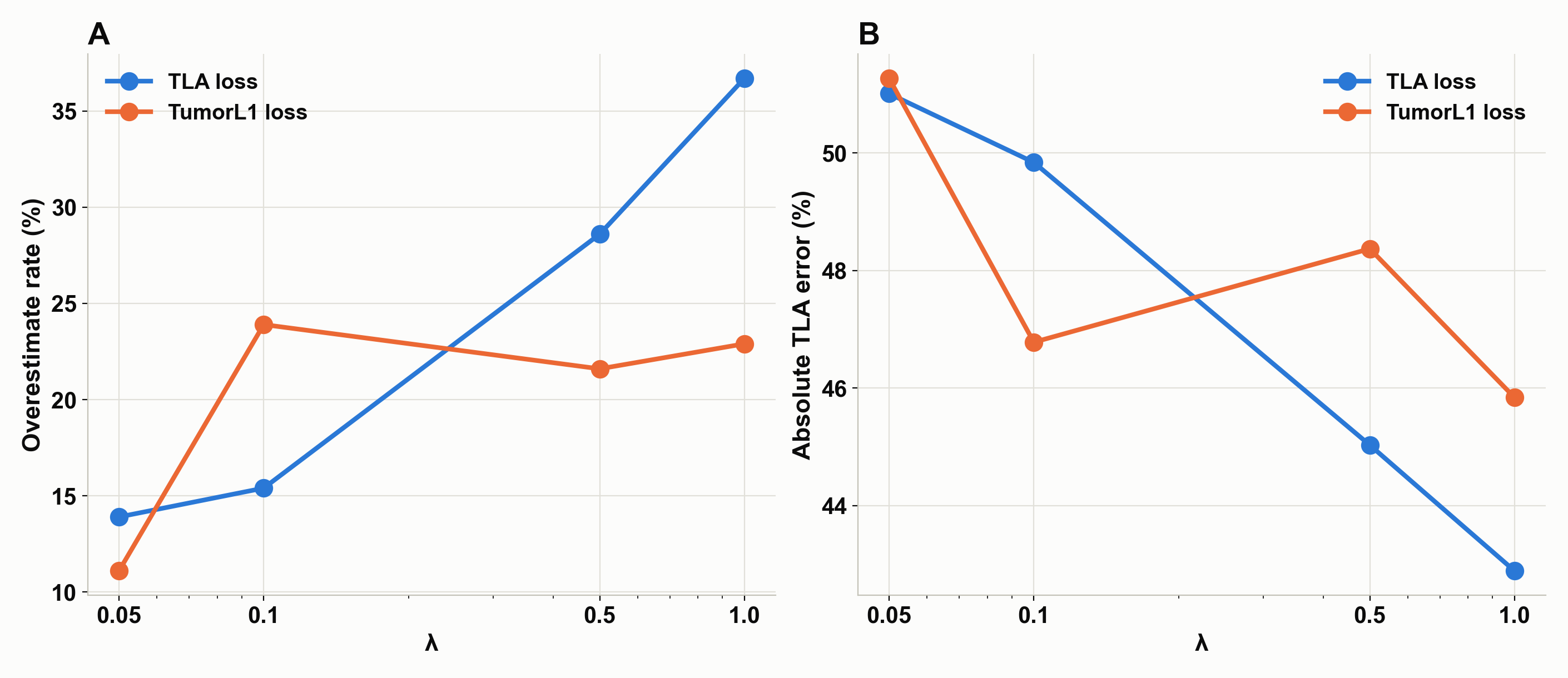}
	\caption{Per-lesion loss formulations across $\lambda \in \{0.05, 0.1, 0.5, 1.0\}$ on $^{18}$F-PSMA. \textbf{(A)} Per-patient overestimate rate. \textbf{(B)} Per-patient absolute TLA error. TumorL1 ($\mathcal{L}_{\text{TumorL1}}$) saturates after $\lambda=0.1$; the TLA loss ($\mathcal{L}_{\text{TLA}}$) continues to respond across the full range.}
	\label{fig:tla_vs_tl1}
\end{figure}

We first compared two formulations of the per-lesion supervision term. The first is the lesion-wise summed-activity loss $\mathcal{L}_{\text{TLA}}$ in Equation~\ref{eq:tla}, which supervises each lesion independently. The second is a uniform L1 loss restricted to tumor-mask voxels, denoted $\mathcal{L}_{\text{TumorL1}}$:

\begin{equation}
\mathcal{L}_{\text{TumorL1}} =
\frac{1}{|\mathcal{M}|}
\sum_{i \in \mathcal{M}}
\left|
\hat{y}_i - y_i
\right|,
\label{eq:tumorl1}
\end{equation}

where $\mathcal{M}$ is the set of tumor-mask voxels, and $\hat{y}_i$ and $y_i$ are the predicted and ground-truth PET values.

Each loss was trained at four weights, $\lambda \in \{0.05, 0.1, 0.5, 1.0\}$, on top of an identical whole-volume L1 base loss, using the 3D U-Net with AFNO bottleneck backbone. Figure~\ref{fig:tla_vs_tl1} shows that $\mathcal{L}_{\text{TLA}}$ and $\mathcal{L}_{\text{TumorL1}}$ respond very differently as $\lambda$ increases. $\mathcal{L}_{\text{TumorL1}}$ improves sharply between $\lambda=0.05$ and $\lambda=0.1$, with absolute TLA error falling from 51.3\% to 46.8\%, and then saturates, fluctuating between 45.8\% and 48.4\% over the remainder of the range, and its overestimation rate flattens near 22--24\%. In contrast, $\mathcal{L}_{\text{TLA}}$ declines steadily across the full range, from 51.0\% to 42.9\%, and its overestimation rate rises correspondingly from 13.9\% to 36.6\%.

Both models show systematic underestimation at low lesion weight. Therefore, the increase in overestimation reflects $\mathcal{L}_{\text{TLA}}$ progressively correcting this bias. $\mathcal{L}_{\text{TumorL1}}$ stops correcting once the tumor-mask error saturates, whereas $\mathcal{L}_{\text{TLA}}$ continues to respond to lesion-level activity mismatch. Moreover, $\mathcal{L}_{\text{TumorL1}}$ penalizes every tumor voxel uniformly, so its gradient is dominated by large lesions and by the bulk of voxels already close to their target. In contrast, the $\mathcal{L}_{\text{TLA}}$'s denominator in Equation~\ref{eq:tla}, ${\log\left(1 + N_k \cdot SUV_{\max}\right)}$, converts the raw size-dependent lesion-activity error into a relative, per-lesion-normalized error. This prevents larger lesions from dominating the objective and helps stabilize gradients across lesions of different sizes. We therefore adopt the lesion-wise $\mathcal{L}_{\text{TLA}}$ formulation. Its higher accuracy at large $\lambda$ comes with an increased overestimation rate, which is why the final model uses a low weight of $\lambda_{\text{TLA}} = 0.05$ and relies on the CT proxy channels and the tumor-contrast loss $\mathcal{L}_c$ to improve tumor-level performance. This keeps the overestimation rate below 20\% in the final model: 19.98\% for $^{18}$F-PSMA with 48.35\% absolute TLA error, and 10.47\% for $^{68}$Ga-PSMA with 64.01\% absolute TLA error, as reported in Table~\ref{tab:tla}.

\begin{table*}[width=1\linewidth,cols=12,pos=h]
\centering
\caption{Cumulative ablation on $^{18}$F-PSMA cohort. Each row adds one component to the row above. ICC values are the mean over features within each class. The shaded row is the full proposed model.}
\label{tab:ablation}
\scriptsize
\setlength{\tabcolsep}{3pt}
\begin{tabular*}{\tblwidth}{@{\extracolsep{\fill}}lccccccccccc@{}}
\toprule
& \multicolumn{4}{c}{Lesion TLA} & \multicolumn{2}{c}{Whole-volume}
& \multicolumn{4}{c}{Tumor-core ICC} \\
\cmidrule(lr){2-5}\cmidrule(lr){6-7}\cmidrule(lr){8-11}
Configuration & $|$err$|$ $\downarrow$ & $\leq$25\% & $\leq$50\% & $\leq$75\%
& SSIM & PSNR & FO & GLCM & GLRLM & GLSZM \\
\midrule
AFNO-L1 (CT only)
& 62.2 & 9.8 & 25.9 & 65.3
& 0.963 & 36.42
& 0.235 & 0.393 & 0.589 & 0.510 \\
\quad + proxies
& 56.5 & 13.0 & 35.3 & 72.9
& \textbf{0.966} & \textbf{37.05}
& 0.328 & 0.532 & 0.723 & 0.481 \\
\quad + $\mathcal{L}_{\text{TLA}}$
& 50.5 & 19.6 & \textbf{51.1} & 78.8
& 0.965 & 36.71
& 0.357 & 0.565 & 0.739 & 0.624 \\
\quad + $\mathcal{L}_{c}$
& \textbf{47.8} & \textbf{25.2} & 54.9 & 80.8
& 0.960 & 35.88
& \textbf{0.392} & \textbf{0.626} & \textbf{0.809} & \textbf{0.760} \\
\rowcolor{lafnocdl}
\quad + $\mathcal{L}_{\text{peri}}$ (LAFNO)
& 48.3 & 24.0 & 50.9 & \textbf{81.0}
& 0.960 & 35.98
& 0.381 & 0.616 & 0.783 & 0.709 \\
\bottomrule
\end{tabular*}
\end{table*}

\subsection{Effect of CT Proxy Conditioning}
\label{sec:abl:proxy}

We first isolated the contribution of the CT-derived proxy channels by training
two models under the same conditions. The first, AFNO-L1, received only the CT
patch as input and was trained with plain whole-volume L1 loss. The second,
AFNO-L1 + Proxy, used the same backbone as AFNO-L1,
but additionally received the CT-derived contrast and disorder proxy channels of
Equations~\ref{eq:contrast}--\ref{eq:disorder}. Since the backbone, decoder, skip connections, and training loss were identical between the two models, this comparison isolates the contribution of CT-derived proxy conditioning.

Adding the proxy channels reduced mean per-patient absolute TLA error from
62.2\% to 56.5\% and increased the fraction of lesions within 50\% of their true
activity from 25.9\% to 35.3\% (Table~\ref{tab:ablation}). Whole-volume image
quality improved only marginally, and radiomics reproducibility improved in
the tumor core. The benefit is therefore concentrated at
the lesion-quantification level rather than distributed uniformly across the
volume. Notably, this arm used only the plain L1 reconstruction loss, with no
tumor-mask, TLA, or peritumoral supervision. This result shows that CT-derived proxy conditioning can improve lesion-level TLA estimation even without lesion segmentations or TLA-specific supervision during training. Unlike the TLA loss, which requires lesion masks to compute ground-truth lesion activity, the contrast and disorder proxies are computed directly from CT. They therefore provide tumor-relevant conditioning by highlighting regions with tumor-like density and texture behavior, even when explicit lesion annotations are unavailable.

\subsection{Effect of the Lesion-wise TLA Loss}
\label{sec:abl:tla}

We next added the lesion-wise TLA term at $\lambda_{\text{TLA}} = 0.05$, holding
the architecture and proxy conditioning fixed. The mean per-patient absolute TLA error reduced from
56.5\% to 50.5\%, and the fraction of lesions within 50\% of true activity rose
from 35.3\% to 51.1\% (Table~\ref{tab:ablation}). Whole-volume fidelity decreased slightly, reflecting the
expected trade between a lesion-focused objective and global reconstruction. The
lesion-level gain substantially outweighs this cost, confirming that explicit
per-lesion activity supervision supplies information the proxy conditioning alone
does not.

\subsection{Effect of the Tumor-Contrast Loss}
\label{sec:abl:contrast}

We next added the tumor-contrast term $\mathcal{L}_c$. This improved lesion quantification
further, reducing absolute TLA error to 47.8\% and raising the fraction of
lesions within 25\% of true activity from 19.6\% to 25.2\%. Its clearest effect,
however, was on intra-lesion structure: tumor-core radiomics reproducibility rose
in every feature class, with the largest gain in GLSZM (0.624 to 0.760),
indicating better recovery of zone-size distribution within the lesion (Table~\ref{tab:ablation}). This is
consistent with $\mathcal{L}_c$ supervising local density structure inside the
tumor mask, which the TLA term, constraining only total activity, does not.

\subsection{Effect of the Peritumoral Loss}
\label{sec:abl:peri}

\begin{figure*}
	\centering
	\includegraphics[width=\textwidth]{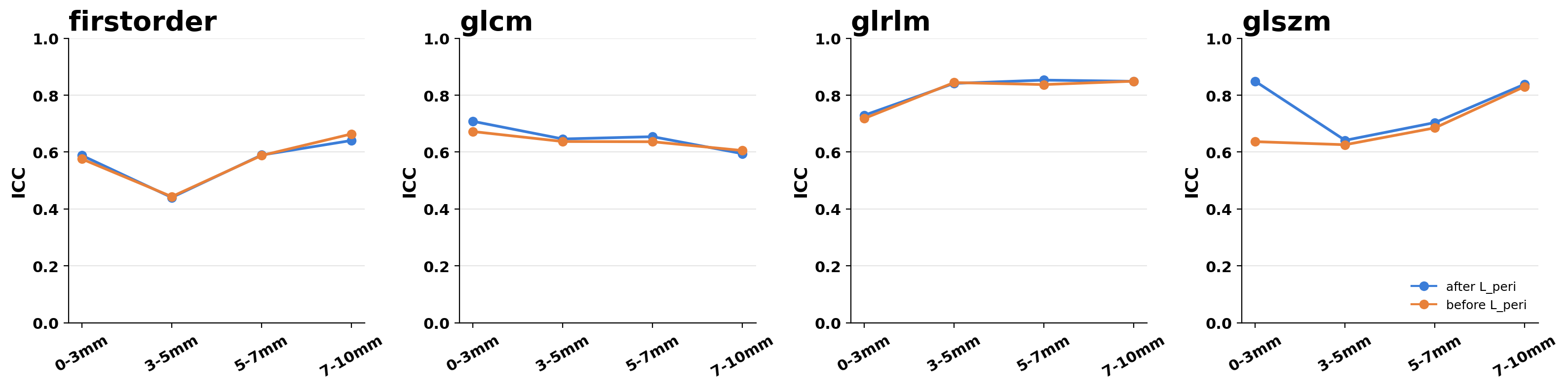}
	\caption{Peritumoral radiomics reproducibility (ICC) by distance band, before and after adding $\mathcal{L}_{\text{peri}}$, for each feature class ($^{18}$F-PSMA). The before/after curves overlap for first-order, GLCM, and GLRLM; the main effect is on GLSZM in the 0--3\,mm band.}
	\label{fig:peri_bands}
\end{figure*}

Finally, we added the peritumoral term $\mathcal{L}_{\text{peri}}$ at
$\lambda_{\text{peri}} = 0.05$, kept deliberately small so that supervision of
the surrounding band does not overpower the tumor-core terms, while leaving
lesion-level and tumor-core metrics essentially unchanged
(Table~\ref{tab:ablation}).

Resolving the effect by feature class and distance band
(Figure~\ref{fig:peri_bands}) shows that the term acts narrowly rather than
broadly. For first-order, GLCM, and GLRLM, peritumoral reproducibility is
almost identical before and after adding $\mathcal{L}_{\text{peri}}$, with the
before/after curves separated by at most 0.02--0.03 ICC at any band. The one
substantial effect is on GLSZM at the immediate rim: size-zone reproducibility
in the 0--3\,mm band rises from 0.637 to 0.849 ($\Delta = 0.21$), while the
same class changes little in the outer bands ($\leq 0.02$). The improvement is
therefore concentrated exactly where the exponential weight is largest,
consistent with the loss design, but it refines a single texture property at
the tumor boundary rather than improving peritumoral fidelity generally.

Peritumoral fidelity nonetheless remains the most challenging region for the
model. The narrowness of this effect likely reflects the small weight assigned
to $\mathcal{L}_{\text{peri}}$, and we did observe a slight drop in tumor-core
ICC, once this term was included, which is expected when balancing many
competing objectives in a single loss. A more carefully designed objective that
jointly penalizes the tumor core and the peritumoral region while establishing
a balance between them may be required to strengthen peritumoral fidelity
without compromising the core.

\section{Discussion}
\label{sec:discussion}

\begin{figure*}
	\centering
	\includegraphics[width=\textwidth]{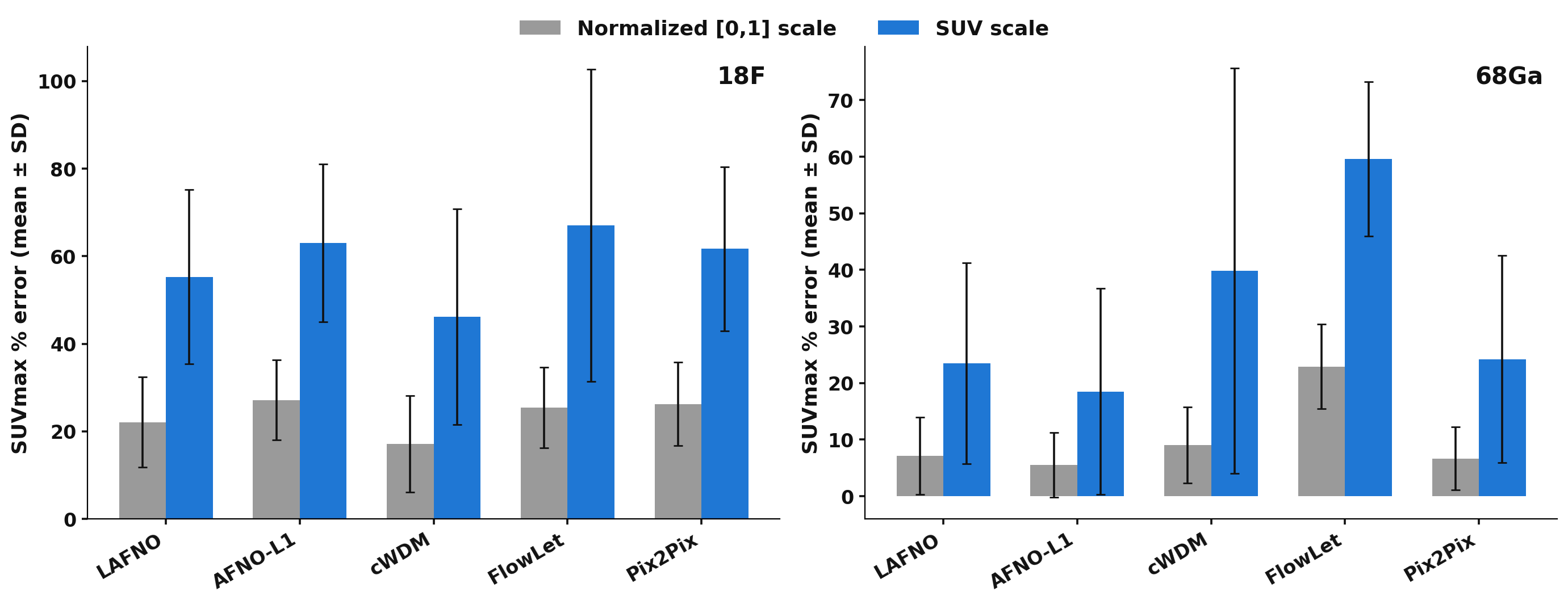}
	\caption{Whole-volume SUVmax percentage error on the normalized $[0,1]$ scale and after conversion back to SUV space, for $^{18}$F-PSMA (left) and $^{68}$Ga-PSMA (right). The inverse log transform amplifies error at the high-uptake end, so SUV-scale error is consistently larger than normalized-scale error across all models.}
	\label{fig:normvssuv}
\end{figure*}

In this study, we show that a biologically conditioned CT-to-PET synthesis model better preserves clinically relevant PET signals than models optimized primarily for global image similarity. LAFNO was designed as a simple AFNO-based synthesis model with CT-derived proxy conditioning and lesion-aware supervision. The AFNO bottleneck was selected because Fourier-domain operators can model long-range spatial interactions and have shown strong performance in medical image translation tasks \cite{bhaskara2025enhancing}. Overall, LAFNO achieved competitive whole-volume image quality while improving lesion-level accuracy and tumor-core radiomics reproducibility.

The whole-volume SSIM, PSNR, and MAE results indicate that LAFNO generates synthetic PET volumes that remain visually close to the ground-truth scans. However, whole-volume SUVmax and SUVmean errors remained relatively high. This is partly expected because the model was trained in log-normalized PET space to handle the large dynamic range of standardized uptake value (SUV) values. When predictions are converted back to SUV space, small errors in log space can become larger absolute errors after inversion, especially in high-uptake regions (Figure~\ref{fig:normvssuv}). This highlights an important limitation of evaluating synthetic PET only using global image metrics or whole-volume SUV errors.

\begin{figure}
	\centering
	\includegraphics[width=.75\textwidth]{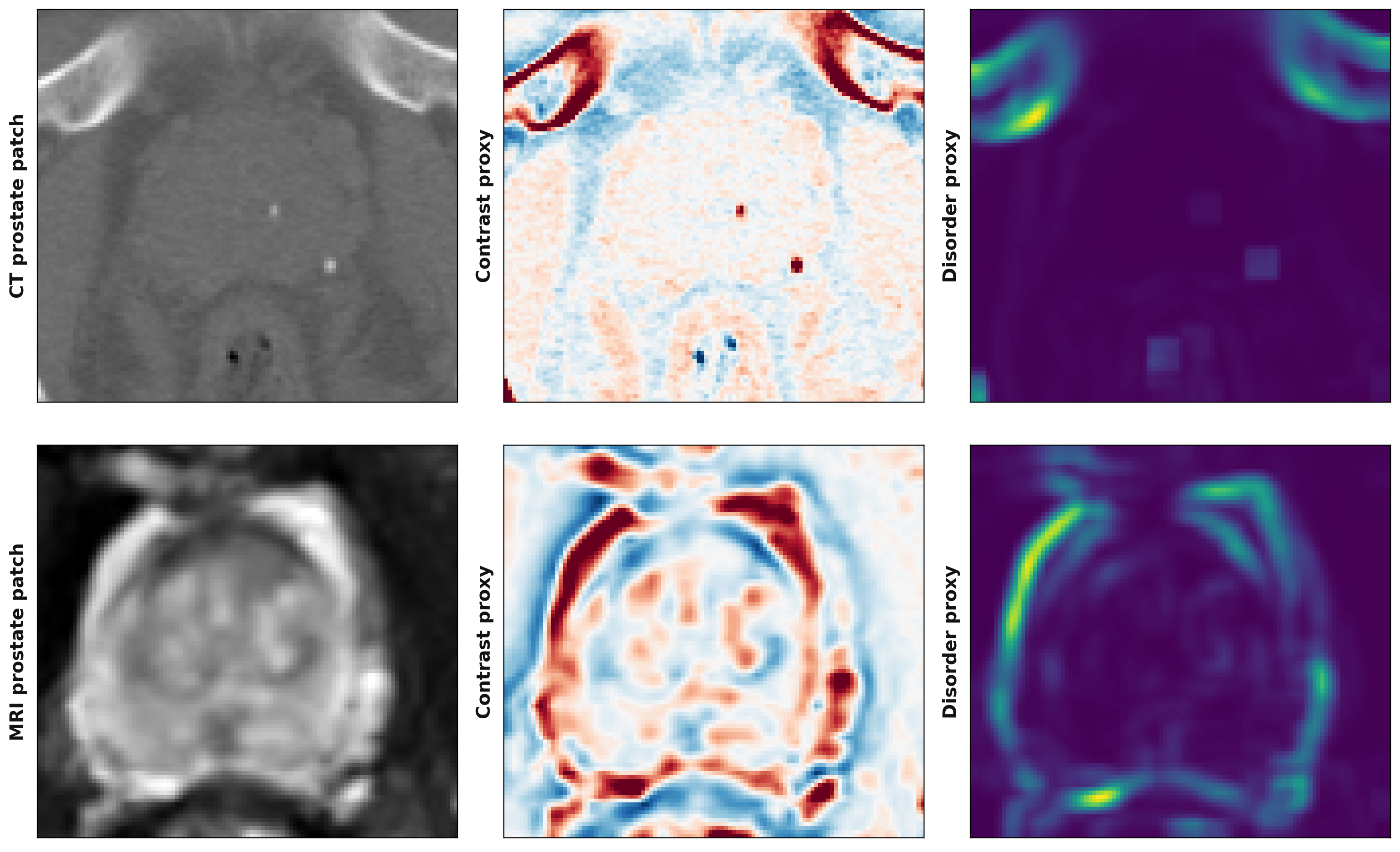}
	\caption{CT- and T2-weighted MRI-derived proxy maps in a prostate-region example. The same contrast and disorder operations highlight a stronger local prostate structure on MRI than on CT.}
	\label{fig:ct_mri_proxy}
\end{figure}

At the lesion level, LAFNO achieved the best TLA performance among the evaluated models, but the absolute TLA error remained high. We attribute part of this residual error to tumor heterogeneity. Tumor lesions are highly heterogeneous (a feature that can challenge the reliability of localized assessments) \cite{almutairi2025immunopet}, and tumor phenotype can vary substantially across lesions and within the same patient \cite{heppner1981tumor}. Similar CT appearances may therefore correspond to different PET uptake patterns depending on lesion biology, tissue state, and tracer avidity \cite{mahdi2026ct}. This is difficult to capture with reconstruction losses dominated by background-region accuracy, which may generate visually plausible PET images while still underestimating or misrepresenting lesion activity \cite{abtahi2026fine}. It is also worth noting that SUV itself is a semiquantitative measure influenced by scanner performance, reconstruction algorithms, calibration, and observer variability, as well as histologic type, tumor grade, and lesion size factors that add further variability independent of model performance \cite{weiss2012suv,ulaner2019fundamentals,wang2025dynamicfdg}. This concern is directly relevant to our dataset, which combines studies acquired across three PET/CT scanners and two PET tracers ($^{18}$F-PSMA and $^{68}$Ga-PSMA) with differing reconstruction parameters variations that can affect measured SUV by more than 50\% according to EANM quality-control guidelines \cite{boellaard2010eanm}. The contrast and disorder proxies were designed as simple CT-derived representations of tumor-relevant density variation and local texture heterogeneity. These proxies do not fully resolve lesion heterogeneity, but they only provide the model with additional spatial cues beyond CT intensity alone. This helps explain why LAFNO improved lesion-level TLA performance compared with the baseline models, even though absolute errors remained substantial. In addition, the effectiveness of proxy conditioning may also depend on the imaging modality. Figure~\ref{fig:ct_mri_proxy} shows that applying the same contrast and disorder operations to a T2-weighted MRI prostate patch highlights stronger local prostate texture information than CT. This suggests that MRI-guided or hybrid CT/MRI proxy conditioning may be better suited for PSMA-PET synthesis in soft-tissue prostate disease.

Similarly, the ICC scores indicated moderate radiomics reproducibility within the tumor core, which is because the tumor interior is also highly heterogeneous. We also observed that LAFNO performed differently across tracers. For $^{18}$F-PSMA, LAFNO improved texture-based ICC values across feature classes, whereas for $^{68}$Ga-PSMA, ICC values were generally lower, and several feature classes remained near 0.5. A similar tracer-dependent pattern was observed in the peritumoral region. Although the peritumoral loss used an exponential distance-decay weighting scheme, recovery near the tumor boundary remained challenging, and for $^{68}$Ga-PSMA, reproducibility peaked in the 3--5\,mm band rather than at the immediate boundary. This behavior may be related to differences in tracer uptake, image reconstruction, noise characteristics, or tumor boundary contrast. For example, $^{68}$Ga-PSMA has a higher average positron energy than $^{18}$F-PSMA, resulting in a longer positron range, reduced spatial resolution, and greater partial volume effects. These factors can degrade tracer quantification and introduce lesion localization uncertainty, which may contribute to reduced model performance. Reconstruction differences may also affect feature reproducibility. In this study, images were acquired across different scanners with various time of flight corrections introducing additional variability that may have influenced model performance. Future studies should aim to minimize inter-tracer and inter-scanner differences by implementing partial volume correction to improve tracer quantification, evaluating anatomically relevant non-PSMA imaging targets that may improve tissue contrast, and standardizing image acquisition, reconstruction, and data preparation workflows. Additional correction for distance dependent peritumoral effects may further improve model robustness.

Finally, the ablation study helps explain how each component contributed to these results. CT-derived proxy conditioning improved lesion-level TLA estimation even without lesion-specific supervision, indicating that the contrast and disorder proxies provide useful tumor-relevant spatial cues. Adding $\mathcal{L}_{\text{TLA}}$ further improved lesion activity preservation by directly supervising each connected component. The tumor-contrast loss $\mathcal{L}_{c}$ produced the clearest gain in tumor-core radiomics reproducibility, suggesting that local contrast supervision helps preserve intra-lesion structure. The peritumoral loss $\mathcal{L}_{\text{peri}}$ had a more localized effect, mainly improving GLSZM reproducibility near the immediate tumor boundary, while leaving most lesion-level and tumor-core metrics largely unchanged.

Several limitations should be noted. The dataset is likely biased toward visually apparent and higher-burden lesions, particularly bone-dominant metastatic disease, because these lesions are easier to identify and annotate on PSMA-PET/CT. As a result, the proxies may not fully represent subtle soft-tissue lesions, small nodal disease, or lesions with weak uptake. Future works will likely require more richly annotated datasets that distinguish lesion phenotype, tissue state, and imaging context. Necrotic, sclerotic, soft-tissue, inflammatory, and physiologic uptake regions may show different PET behavior even when their CT appearance overlaps. Distinguishing these tissue types and characterizing how the underlying biological signal varies across them would allow models to learn subtype-specific patterns rather than relying on a single generalized CT-to-PET mapping. Such annotations could also help identify more specific CT and PET signatures for different lesion groups and guide the design of future conditioning strategies. Another major limitation is that high physiologic bladder activity can introduce spillover, noise, and boundary-related uncertainty in the pelvic region, which may affect both training and evaluation near the prostate bed and adjacent soft tissues. Future datasets could reduce this source of uncertainty by using acquisition protocols that limit bladder activity, such as catheter-assisted bladder drainage or other standardized bladder-management strategies during image acquisition. Models trained on such datasets may better learn prostate-bed activity with reduced spillover from high bladder uptake.

\section{Conclusion}

In this work, we proposed LAFNO for CT-to-PSMA-PET synthesis using CT-derived proxy conditioning and lesion-aware supervision. The model was designed to preserve clinically relevant tumor information. LAFNO also maintained a competitive global image quality while improving lesion-level activity preservation and tumor-core radiomics reproducibility compared with the baseline models. However, peritumoral reproducibility and tracer-dependent differences, especially for $^{68}$Ga-PSMA, remain challenging. Future work should focus on larger annotated datasets, stronger biological conditioning, and improved modeling of tumor-adjacent tissue.













\FloatBarrier
\bibliographystyle{cas-model2-names}

\bibliography{cas-refs}



\end{document}